%% file: egpaper_for_review.tex
\documentclass[10pt,twocolumn,letterpaper]{article}

\usepackage{cvpr}
\usepackage{times}
\usepackage{epsfig}
\usepackage{graphicx}
\usepackage{amsmath}
\usepackage{amssymb}

\usepackage[T1]{fontenc}
\usepackage{subfigure}
\usepackage[lined,ruled,linesnumbered]{algorithm2e}

\PassOptionsToPackage{hyphens}{url}
\usepackage[pagebackref=true,breaklinks=true,letterpaper=true,colorlinks,bookmarks=false]{hyperref}

\cvprfinalcopy 

\def\cvprPaperID{\vspace{-0.5cm}} 

\ifcvprfinal\fi
\begin{document}

\title{Memory Based Online Learning of Deep Representations from Video Streams\vspace{-0.5cm}}
\author{Federico Pernici, Federico Bartoli, Matteo Bruni and Alberto Del Bimbo
\\
MICC -- Media Integration and Communcation Center\\
University Of Florence\\
}

\maketitle

\begin{abstract}
We present a novel online unsupervised method for face identity learning from video streams. The method exploits deep face descriptors together with a memory based learning mechanism that takes advantage of the temporal coherence of visual data. 
Specifically, we introduce a discriminative feature matching solution based on Reverse Nearest Neighbour and a feature forgetting strategy that detect redundant features and discard them appropriately while time progresses. 
It is shown that the proposed learning procedure is asymptotically stable and can be effectively used in relevant applications like multiple face identification and tracking from unconstrained video streams. Experimental results show that the proposed method achieves comparable results in the task of multiple face tracking and better performance in face identification with offline approaches exploiting future information. Code will be publicly available.
\vspace{-0.5cm}
\end{abstract}
\section{Introduction}
\label{sec:introduction}
Visual data is massive and is growing faster than our ability to store and index it, nurtured by the diffusion and widespread use of social platforms. Their fundamental role in advancing object representation, object recognition and scene classification research have been undoubtedly assessed by the achievements of Deep Learning \cite{krizhevsky2012imagenet}. However, the cost of supervision remains the most critical factor for the applicability of such learning methods as linear improvements in performance require an exponential  number of labelled examples \cite{sun2017revisiting}.  Efforts to collect large quantities of annotated images, such as ImageNet \cite{deng2009imagenet} and Microsoft coco \cite{lin2014microsoft}, 
while having an important role in advancing object recognition, don't have the necessary scalability and are hard to be extended, replicated or improved. They may also impose a ceiling on the performance of systems trained in this manner. Semi or unsupervised Deep Learning from image data still remains hard to achieve.
\begin{figure}[t]
\centering	
\includegraphics[width=1.03\columnwidth]{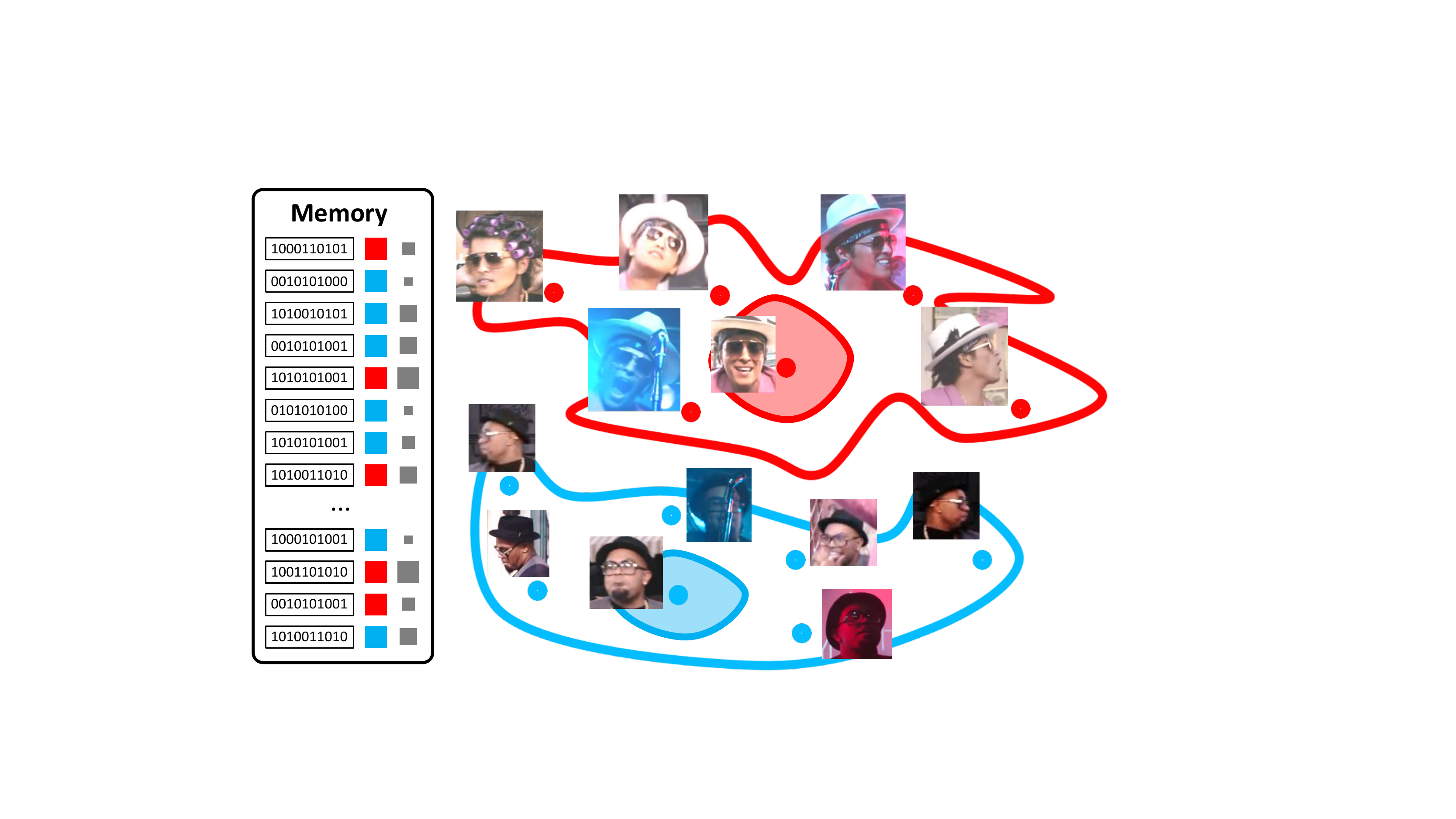}
\caption{Memory based appearance representation. \emph{Left}: Each element in the memory consists of a descriptor with an associated identity (indicated by box color) and an associated scalar value reflecting the degree of redundancy (indicated by grey box area) with respect to the current representation. \emph{Right}: The shaded regions represent the original appearance representation (i.e. VGGface). The descriptors outside those regions are learned from the video and extend the original appearance representation.
\vspace{-0.6cm}
}
\label{intro}
\end{figure}
\\
\indent
An attracting alternative would be to learn the object appearance from video streams with no supervision, both exploiting the large quantity of video available in the Internet and the fact that adjacent video frames contain semantically similar information. This provides a variety of conditions in which an object can be framed, and therefore a comprehensive representation of its appearance can be obtained.
Accordingly, tracking a subject in the video could, at least in principle, support a sort of unsupervised incremental learning of its appearance. This would avoid or reduce the cost of annotation as time itself would provide a form of self-supervision.  However, this solution is not exempt of problems \cite{wired}. On the one hand, parameter re-learning of Deep Networks, to adequately incorporate the new information without catastrophic interference, is still an open challenge \cite{li2016learning,rusu2016progressive}, especially when re-learning should be done in real time while tracking, without the availability of labels and with data coming from a stream which is often non-stationary.
On the other hand, classic object tracking \cite{Kristan_2017_ICCV} has substantially divergent goals from continuous incremental learning. While in tracking the object appearance is learned only for detecting the object in the next frame (the past information is gradually \emph{forgotten}), continuous incremental learning would require that \emph{all} the past visual information of the object observed so far is collected in a comprehensive and cumulative representation. This requires that tracking does not drift in the presence of occlusions or appearance changes and that incremental learning should be asymptotically stable in order to converge to an univocal representation.
\\
\indent
In this paper, we present a novel online unsupervised method for face identity learning from unconstrained video streams. The method exploits CNN based face detectors and descriptors together with a novel incremental memory based learning mechanism that collects descriptors and distills them based on their redundancy with respect to the current representation. This allows building a sufficiently compact and complete appearance representation of the individual identities as time advances (Fig.~\ref{intro}). 
\\
\indent
While we obtained comparable results with offline approaches exploiting future information in the task of multiple face tracking, our model is able to achieve better performance in face identification from unconstrained video. In addition to this, it is shown that the proposed learning procedure is asymptotically stable and the experimental evaluation confirms the theoretical result.
In the following, in Section ~\ref{RelatedWork}, we cite a few works that have been of inspiration for our work. In Section ~\ref{Contributions} we highlight our contributions, in Section ~\ref{approach} we expounded the approach in detail and finally, in Section ~\ref{Experiments}, experimental results are given.
\section{Related Work}
\label{RelatedWork}
\textbf{Memory Based Learning:} Inclusion of a memory mechanism in learning \cite{kumaran2016learning} is a key feature of our approach. On domains that have temporal coherence like Reinforcement Learning (RL) memory is used to store the past experience with some priority and to sample mini-batches to perform incremental learning \cite{mnih2015human} \cite{Schaul2016}. This makes it possible to break the temporal correlations by mixing more and less recent experiences. 
More recently, Neural Turing Machine architectures have been proposed in \cite{graves2014neural,graves2016hybrid} and  \cite{santoro2016meta} that implement an augmented memory to quickly encode and retrieve new information. These architectures have the ability to rapidly bind never-before-seen information after a single presentation via an external memory module. However, in these cases, training data are still provided supervisedly and the methods don't scale with massive video streams. 
\\
\indent
\textbf{Open Set:} In addition to the incremental learning procedure, the system needs to have the capability to discriminate between already known and unknown classes (\emph{open set}) \cite{scheirer2013toward}. 
The open set classification is a problem of balancing known space (specialization) and unknown open space (generalization) according to the class rejection option.
Formalization for open space risk is considered as the relative measure of open space compared to the overall measure space \cite{scheirer2013toward,Scheirer_2014_TPAMIb,Bendale_2016_CVPR,rudd2017extreme}. 
The underlying assumption in these approaches is that data is \textsc{i.i.d.} which allows sampling the overall space uniformly. However, in a continuously data stream context, as in this paper, data is no longer independent and identically distributed, therefore balancing the known space vs the unknown space is more difficult since space with absence of data may be misinterpreted for open space. Storing data in a memory module can limit these effects
\cite{cornuejols2006,cornuejols2010}.
\\
\indent
\textbf{Open World:} The other fundamental problem is incorporating the identified novel classes into the learning system (\emph{open world}) \cite{bendale2015towards}. 
This requirement favors non-parametric methods, since they can quickly learn never seen before information by simply storing examples. The Nearest Class Mean (NCM) classifier proposed in \cite{mensink2013distance}, has been shown to work well and be more robust than standard parametric classifiers in an incremental learning setting \cite{mensink2013distance} \cite{mensink2012metric} \cite{ristin2014incremental}.
NCM's main shortcomings are: it is not a discriminative classifiers and nonlinear data representation and/or non \textsc{i.i.d.} data streams limit the effectiveness of using the mean. We adopt from NCM the idea of prototype-based classification. However, the prototypes we use are not the average features vectors but we keep a representative non redundant discriminative subset.
\\
\indent
\textbf{Multiple Object Tracking:} All the methods we described so far make use of ground truth labels and typically address the categorization problem in which data is manually cropped around the object boundary.
An alternative approach that in principle accomplishes the class-incremental learning criteria expounded above (i.e. \emph{open set} and \emph{open world}) but with the addition of unknown labels and with data coming from the output of a detector (i.e. no manual cropped data) is Multiple Object Tracking (MOT) \cite{leal2017tracking,surveyMot}. Recent Multiple Object Tracking algorithms typically adopt appearance and motion cues into an affinity model to estimate and link detections to form tracklets which are afterwards combined into final trajectories \cite{brendel2011multiobject, zhang2008global, andriyenko2012discrete, li2006robust, li2008tracking,bae2014robust}. 
Most existing MOT methods are applied to pedestrian tracking and either use simple color histogram features \cite{zhang2008global,shitrit2011tracking,huang2008robust,
li2009learning,wu2013simultaneous} or hand-crafted features \cite{roth2012robust,wang2017tracklet,kuo2011does,cinbis2011unsupervised} as the appearance representation of objects and have simple appearance update mechanisms.
Few exceptions can operate online and use deep features \cite{yu2016poi,Bewley2016_sort,sadeghian2017tracking,WojkeBP17} but they still assume continuous motion and do not update the appearance. MOT methods are not suited to abrupt changes across different shots or scenes since the assumptions of continuous motion no longer hold.
Abrupt changes across different shots are typically handled offline by exploiting tracklets into predetermined non-overlapping shots as in clustering face descriptors \cite{wu2013constrained} \cite{tapaswi2014total} \cite{xiao2014weighted} \cite{zhang2016tracking}.
\\
\indent
\textbf{Long Term Object Tracking:} Finally, another relevant research subject to our learning setting is long-term object tracking \cite{LTDT2014}. The aim of long-term object tracking is to track a specific object over time and re-detect it when the target leaves and re-enters the scene. Only a few works on tracking have reported drift-free results on on very long video sequences (\cite{kalalCVPR2010, MedioniCVPR2011, alienpami, hua2014occlusion, Hong_2015_CVPR} among the few), and only few of them have provided convincing evidence on the possibility of incremental appearance learning strategies that are asymptotically stable \cite{kalalCVPR2010}\cite{alienpami}. However, all of these works only address tracking and perform incremental learning to detect the target in the next frame. 
\\
\vspace{-0.5cm}
\section{Contributions}
\label{Contributions}
1. We firstly combine in a principled manner Multiple Object Tracking in an online \emph{Open World} learning system in which the learning strategy is shown to be asymptotically stable.

2. The proposed method performs very well with respect to offline clustering methods which exploits future information. 

3. Different from several existing approaches, our proposed method operates online and and hence have a wider range of applications particularly face recognition with auto-enrollment of unrecognized subjects.
\section{The proposed approach }
\label{approach}
\begin{figure}[t]
\centering	
\includegraphics[width=0.99\columnwidth]{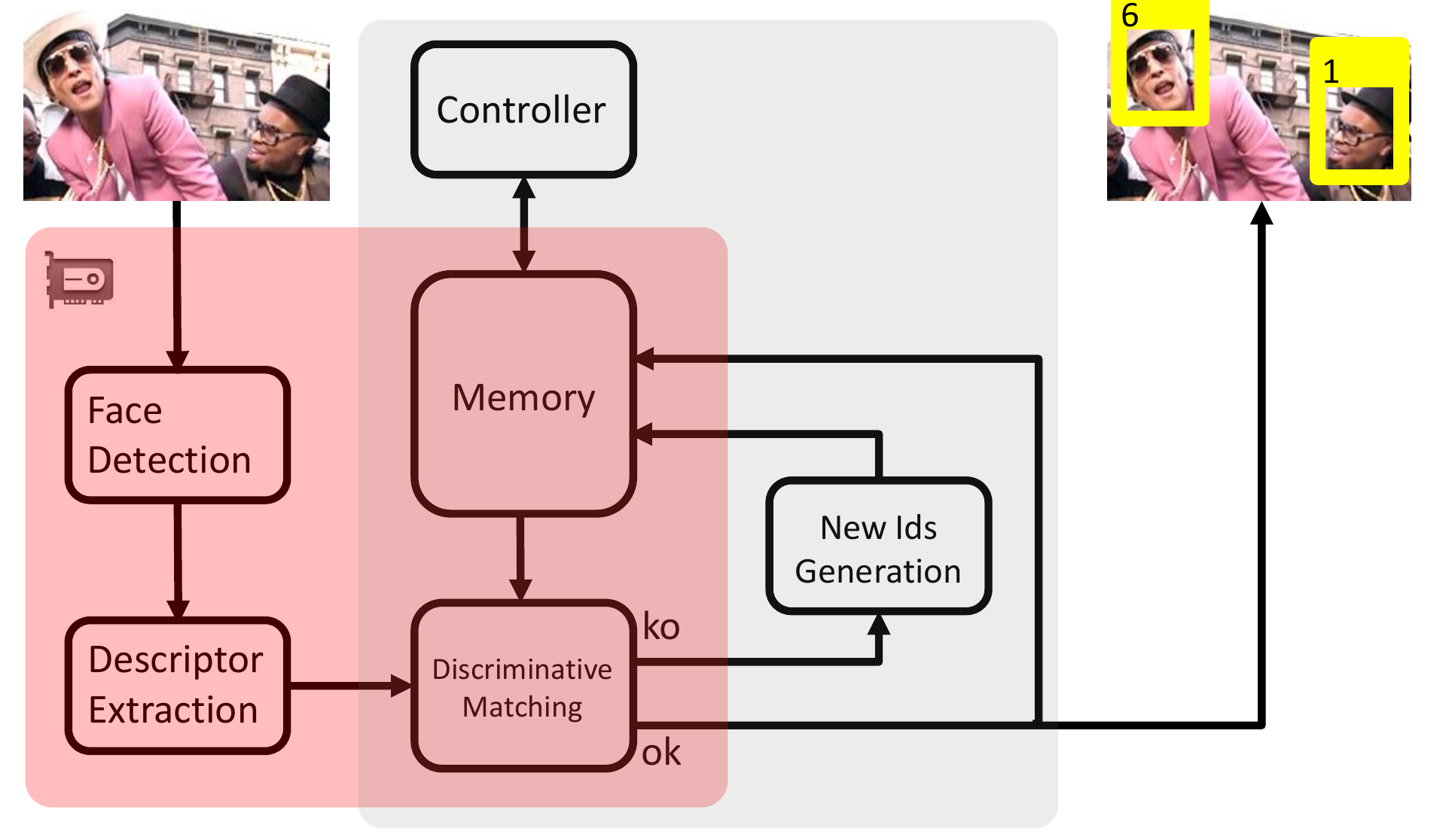}
\caption{Block diagram presenting the major work flow and functional 
components in the proposed method. The gray shaded region highlights
the components discussed in this paper. The memory module and the matching strategy run on the GPU.\vspace{-0.4cm}}
\label{diagram}
\end{figure}
In our system, deep face descriptors are computed on face regions detected by a face detector and stored in a memory module as:
\begin{eqnarray}
\mathcal{M}(t) =  \{  (\mathbf{x}_i,\text{Id}_i, {e}_i, {a}_i)  \}^{N(t)}_{i=1}
\end{eqnarray}
where $\mathbf{x}_i$ is the deep descriptor, $\text{Id}_i$ is the object identity (an incremental number), ${e}_i$ is the eligibility factor (discussed in the following), ${a}_i$ tracks the age of items stored in memory and $N(t)$ is the number of descriptors at time $t$ in the memory module. 
\\
\indent
The block diagram of the proposed system is shown in Fig.~\ref{diagram}.
As video frames are observed, new faces are detected and their descriptors are matched with those already in the memory. Each  newly observed descriptor will be assigned with the object identity of its closest neighbour according to a discriminative strategy based on reverse nearest neighbor described in the next section. Unmatched descriptors of the faces in the incoming frame are stored in the memory module with a new $\text{Id}$. They ideally represent hypothesys of new identities  that have not been observed yet and will eventually appear in the following frames. In order to learn a cumulative and comprehensive identity representation of each observed subject, two distincts problems are addressed. They are concerned with matching in consecutive frames and control of the memory module. These are separately addressed in the following subsections respectively.
\begin{figure}[b]
\centering    
\includegraphics[width=0.99\columnwidth]{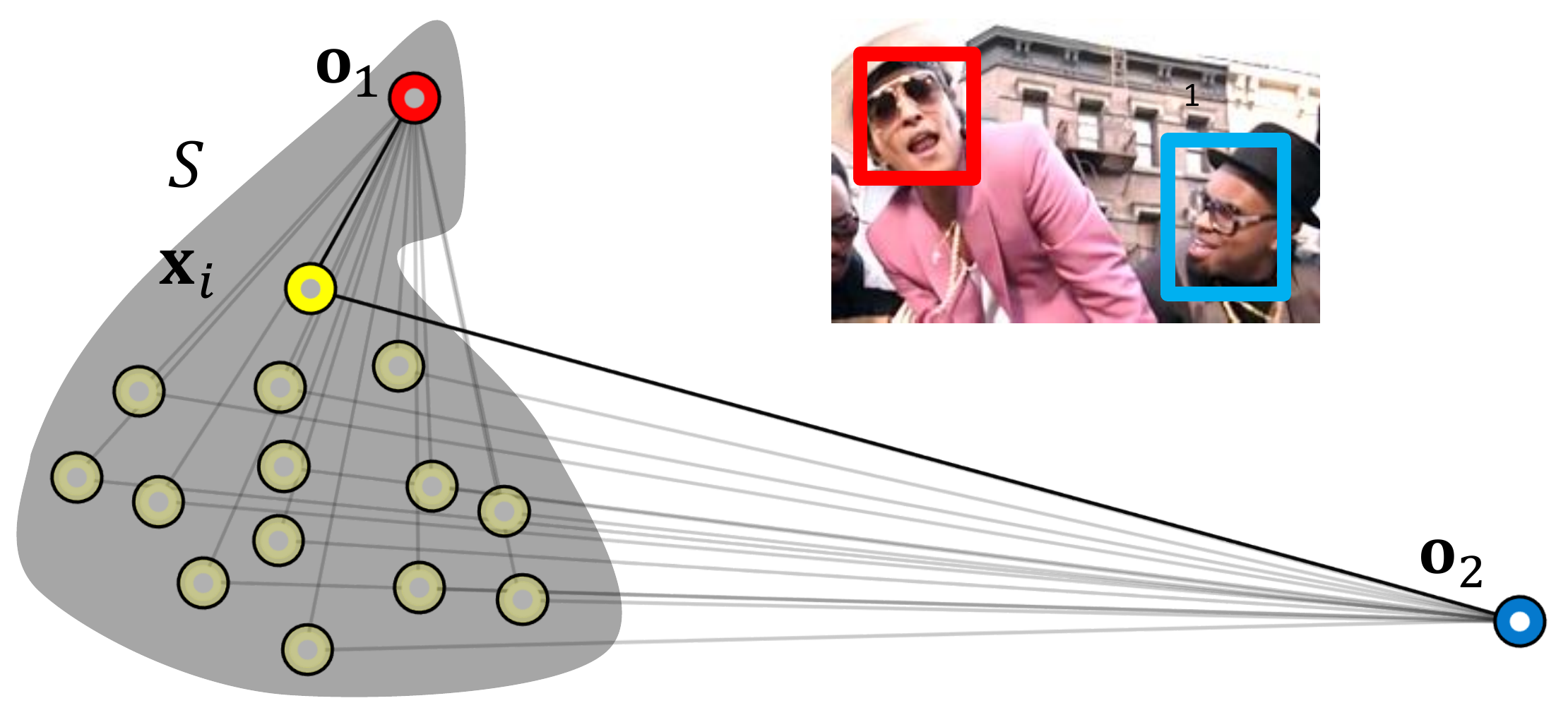}
\caption{Reverse Nearest Neighbor for a repeated temporal visual structure ($S$) with the distance ratio criterion. All elements $\mathbf{x}_i$ match with $\mathbf{o}_1$, for clarity only one of them is highlighted to show the distances (thick black lines). }
\label{fig_RNN}
\end{figure}
\vspace{-0.4cm}
\subsection{Reverse Nearest Neighbour Matching} 
\label{Template  updating}
While tracking in consecutive frames, it is likely that the face of the same individual will have little differences from one frame to the following. 
In this case, highly similar descriptors will be stored in the memory and quickly a new face descriptor of the same individual will have comparable distances to the nearest and the second nearest descriptor already in the memory. 
In this case, a discriminative classifier like the Nearest Neighbor (NN) based on the distance-ratio criterion \cite{lowe} does not work properly and matching cannot be assessed. We solved this problem  by performing descriptor matching according to Reverse Nearest Neighbour (ReNN) \cite{Korn2000}:  
\begin{eqnarray}
\resizebox{.97\hsize}{!}{
$
\mathcal{M}^\star = \Big \{  (\mathbf{x}_i,\text{Id}_i, e_i, a_i) \in \mathcal{M}(t) \: | \: \frac{||\mathbf{x}_i-1\mathrm{NN}_{I_t}(\mathbf{x}_i)||}{||\mathbf{x}_i-2\mathrm{NN}_{I_t}(\mathbf{x}_i)||} < \bar{\rho}, \, \Big  \}
$
}
\label{RNNtemplate}
\end{eqnarray}
where $\bar{\rho}$ is the distance ratio threshold for accepting a match, $\mathbf{x}_i$ is a deep face descriptor in the memory module and $1\mathrm{NN}_{I_t}(\mathbf{x}_i)$ and $2\mathrm{NN}_{I_t}(\mathbf{x}_i)$ are respectively its nearest and second nearest neighbor deep face descriptor in the incoming frame $I_t$.
\\
\indent
Fig.~\ref{fig_RNN} shows the effects of this change of perspective: here two new observations are detected (two distinct faces, respectively marked as $\mathbf{o}_1$ and $\mathbf{o}_2$). They both have distance ratio close to 1 to the nearest $\mathbf{x}_{i}$s in the memory (the dots inside the grey region $S$). Therefore both their matchings are undecidable. Differently from NN, ReNN is able to correctly determine the nearest descriptor for each new descriptor in the incoming frame. In fact, with ReNN, the roles of $\mathbf{x}_i$ and $\mathbf{o}_i$ are exchanged and the distance ratio is computed between each $\mathbf{x}_i$ and the $\mathbf{o}_{i}$ as shown in figure for one of the $\mathbf{x}_i$s (the yellow dot is associated to the newly observed red dot). 
Due to the fact that with ReNN a large number of descriptors (those accumulated in the memory module) is matched against a relatively small set of descriptors (those observed in the current image), calculation of the ratio between distances could be computationally expensive if sorting is applied to the entire set. However, minimum distances can be efficiently obtained by performing twice a brute force search, with parallel implementation on GPU \cite{li2015brute}. This technique not only leverages the very efficient CUDA matrix multiplication kernel for computing the squared distance matrix but it also exploits the GPU parallelism since each query is independent. GPU limited bandwidth is not an issue being the memory incrementally populated.
\\
\indent
The other important advantage of using ReNN is that all the descriptors $\mathbf{x}_i$ of the shown repeated structure $S$ of Fig.~\ref{fig_RNN} match with the descriptor $\mathbf{o}_1$ resulting in a one to many correspondence: 
$\{ \mathbf{o}_1 \} \leftrightarrow  \{ \mathbf{x}_i \}$.
This capability provides a simple and sound method in the selection of those redundant descriptors that need to be condensed into a more compact representation. The feature $\mathbf{o}_1$ will be used, as described in the next section, to influence the other matched (redundant) features $\mathbf{x}_i$ regarding the fact that they belong to the same repeated structure. 
Therefore not only ReNN restores the discriminative matching capability under the distance ratio criterion but it also creates the foundation for the development of memory control strategies to correctly forget the redundant 	feature information.
\subsection{Memory Control} 
Descriptors that have been matched according to ReNN ideally represent different appearances of a same subject face. However, collecting these descriptors indefinitely could quickly determine memory overload. To detect redundant descriptors and discard them appropriately, we defined a dimensionless quantity $e_i$ referred to as \emph{eligibility}. This is set to $e_i = 1$ as a descriptor is entered in the memory module and hence decreased at each match with a newly observed descriptor, proportionally to the distance ratio: 
\begin{equation}
e_{i}(t+1) = \eta_{i} \, e_{i}(t). 		\label{eligibility}
\end{equation}
When doing this, we also re-set the age: $a_i=0$. 
Eligibility allows to take into account both discriminative spatial redundancy at a rate proportional to the success of matching in consecutive frames. 
In fact, as the eligibility $e_{i}$ of a face descriptor $\mathbf{x}_i$ in the memory drops below a given threshold $\bar{e}$ (that happens after a number of matches), that descriptor with its associated identity, age and relative eligibility is removed from the memory module: 
\begin{equation}
\mbox{if} \: (e_{i} < \bar{e}) \: \mbox{then} \: \mathcal{M}(t+1) = \mathcal{M}(t) \setminus \{ (\mathbf{x}_i,\text{Id}_i, {e}_i, a_i) \}.
\label{eligibilityThreshold}
\end{equation}
The value $\eta_{i}$ is computed according to:
\begin{equation}
	\eta_{i} = \bigg[  \frac{1}{\bar{\rho}}
	\frac{d^{1}_i}	
				{d^{2}_i} \bigg]^\alpha,			
    \label{eq_squaredratio}
\end{equation} 
where $d^{1}_i$ and $d^{2}_i$ are respectively the distances between $\mathbf{x}_i$ and its first and second nearest neighbour $\mathbf{o}_i$, the value $\bar{\rho}$ is the distance-ratio threshold of Eq.~\ref{RNNtemplate} here used to normalize 
$\eta_i$ in the unit interval. The value of $\alpha$ emphasizes the effect of the distance-ratio. 
With every memory update we also increment the age $a_i$ of all non-matched elements by 1.
\begin{figure}[t]
\centering    
\includegraphics[width=0.8\columnwidth]{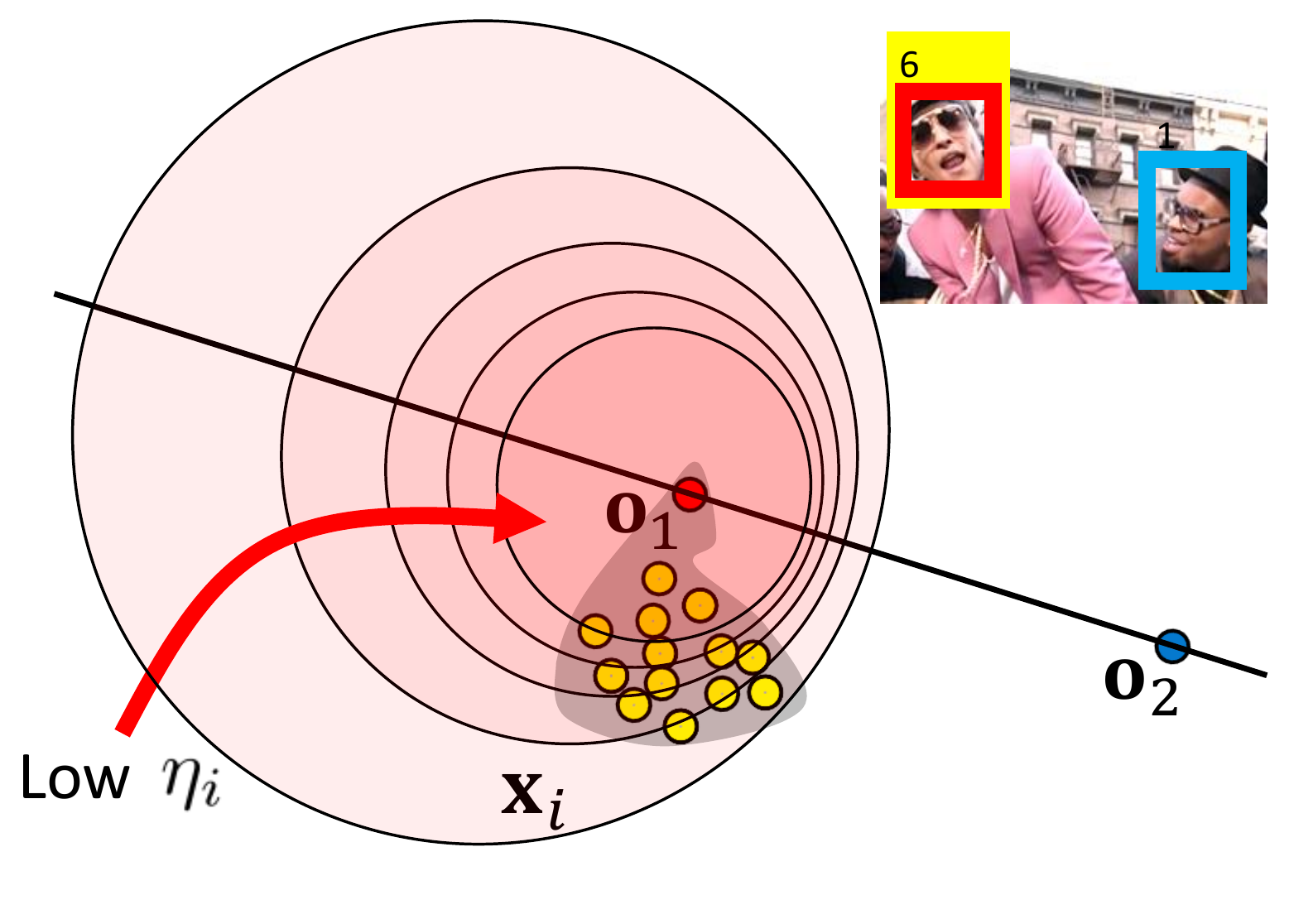}
\caption{The shape of the density (here in 2D) down-weighting the eligibility associated to each matched descriptor in the memory. Features $\mathbf{x}_i$ in proximity of the observed descriptor $\mathbf{o}_1$ have their eligibility decreased to encourage their redundancy. The asymmetric shape of the density encourages more diversity in the \emph{open space} far from the identity $\mathbf{o}_2$ rather than close. }
\label{fig_density}
\vspace{-0.4cm}
\end{figure}
Eq.~\ref{eq_squaredratio} defines a density that weights more the eligibility around the matched features and less the eligibility far apart from their second nearest neighbor. This definition is similar to discriminative distance metric learning in which the features belonging to two different classes are expected to be separated as much as possible in the feature space. The density defined by Eq.~\ref{eq_squaredratio} can be visualized in Fig.~\ref{fig_density} for some values of the distance ratio below the matching threshold $\bar{\rho}$. Each 2D circle in the figure visually represents the density weighting the eligibility of the matching descriptors. 
The geometric shape of the density is a generalization to multiple dimensions of the Apollonious circle\footnote{Apollonius of Perga (c. 262 BC - c. 190 BC) showed that a circle may also be defined as the set of points in a plane having a constant ratio of distances to two fixed foci.}. 
In particular, the asymmetric shape of the density induced by the distance ratio encourages learning feature diversity in the open space. Therefore not only the matching is discriminative and indicated for rejecting hypotheses (\emph{Open Set}) but also well suited for learning in an \emph{Open World}.
\subsection{Temporal Coherence in Image Space}
\label{sec:tempco}
The model previously described exploits video temporal coherence in the deep descriptor space, further spatio-temporal coherence is exploited in the image space introducing the following constraints:

\textbf{1. Id novelty}: Potential novel identities in the current frame are included in the memory only if at least one known identity is recognized in the current frame. This allows introducing novel identity information which is known to be reasonably different from the recognized ones.

\label{idTracking}
\textbf{2. Id temporal coherence}: An identity is assigned and included in the memory only if has been assigned in two consecutive frames. After the assignment (i.e. memory inclusion) it must match at least once in the following 3 frames, otherwise it is discarded.

\textbf{3. Id uniqueness}: Duplicated Ids in the current frame are not considered.

\begin{figure}[t]
\centering	
\includegraphics[width=0.4\textwidth]{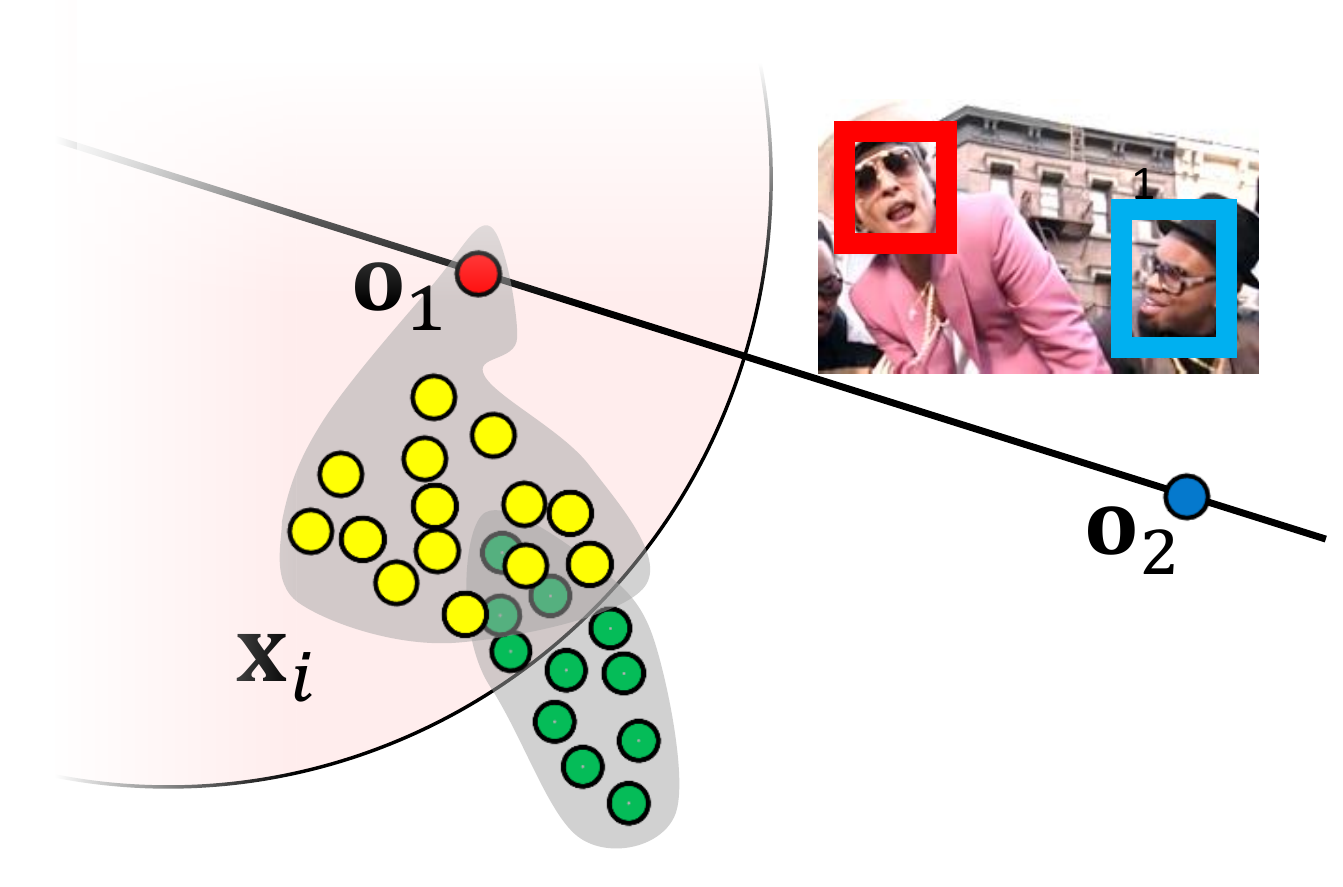}
\caption{Matching with multiple identities. The identity $\mathbf{o}_1$ matches with two identities (yellow and green). The ambiguity is resolved by assigning $\mathbf{o}_1$ with the Id having the largest number of matched descriptors (i.e. the yellow identity).\vspace{-0.4cm}}
\label{Ambiguity}
\end{figure}

\textbf{4. Id ambiguity}: A subject may match with multiple identities. This ambiguity is resolved by assigning all the matched descriptors with the Id having the largest number of matched descriptors as shown in Fig.~\ref{Ambiguity}.

Bounding box overlap, typically used in multiple object tracking, is not exploited since not effective in unconstrained video with abrupt motions. Video temporal coherence in the image space is explicitly enforced by the 2nd constraint. 
\subsection{Memory Overflow Control}
Our method, operating online, does not require any prior information about how many identity classes will occur, and can run for an unlimited amount of time. However, since the memory requirement is dominated by the size of the stored exemplars, if the number of identities increases indefinitely the exemplar removal based on eq.~\ref{eligibilityThreshold} may not be sufficient in handling redundancy and the system may overflow its bounded memory. In this condition the system is forced to remove some stored exemplars by the memory limitations.
To overcome this issue we follow a strategy similar to \cite{santoro2016meta,KaiserNRB17} that involves the use of a policy based on removing from the memory the Least Recently Used Access (LRUA) exemplars. 
This is achieved by finding memory items with maximum age $a_i$ in the memory, and write to one of those. 
Therefore the system preserves recently encoded information according to the Eligibility strategy, or writes to the last used location according to the LRUA strategy. The latter can function as an update of the memory with newer, possibly more relevant information by avoiding the deletion of rare but useful descriptors. A benefit of the LRUA strategy is that of handling those features collected in the memory that will never obtain matches. This effect is largely due to scene occluders or with descriptors extracted from bounding boxes computed from false positives of the face detector. In the long run such features may waste critical space in the memory buffer. 
\subsection{Asymptotic stability}
\label{AsymptoticStability}
\begin{figure}[t]
\centering	
\includegraphics[width=0.42\textwidth]{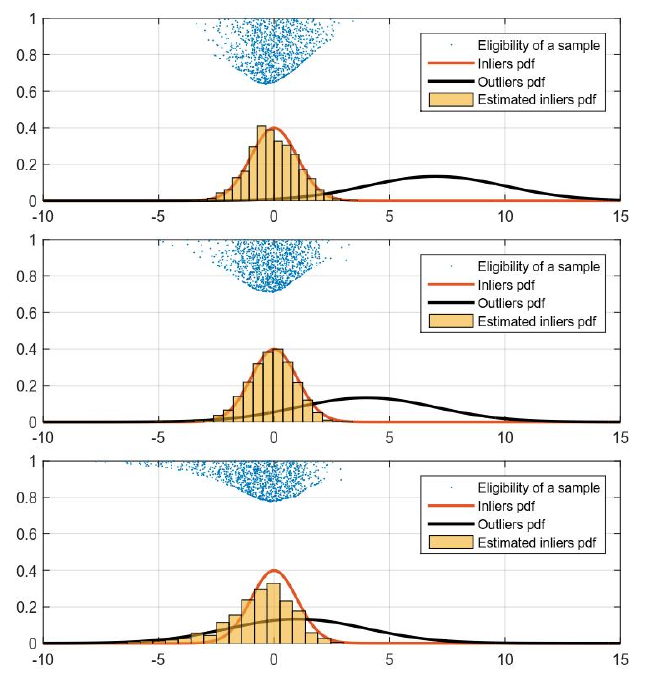}
\caption{Asymptotic stability of incremental learning of a face identity in a sample sequence}\vspace{-0.8cm}.
\label{asymptotic}
\end{figure}

Under the assumption that descriptors are sufficiently distinctive (as in the case of deep face descriptors), the incremental learning procedure described above stabilizes asymptotically around the probability density function of the descriptors of each individual subject face. This can be proved by studying the discrete dynamic system of Eq.~\ref{eligibility} relating $e(t+1)$ to $e(t)$ by the map $T:X \mapsto X$ as $e(t+1)= T(e(t))$. A fixed point of $T$ corresponds to an equilibrium state of the discrete dynamical system. In particular if $T$ is a contraction there is a unique equilibrium state and the system approaches this state as time goes to infinity starting from any initial state. In this case the fixed point is globally asymptotically stable. More formally:
\newtheorem{theorem}{Theorem (Contraction Mapping)}
\begin{theorem}
Let $({X}, d)$ be a complete metric space and  $T: {X} \mapsto {X}$ be the map of Eq.~\ref{eligibility} such that $d( T(e),T(e^\prime) ) \leq c \cdot d(e, e^\prime)$  for some $0 < c \leq 1$ and all $e$ and $e^\prime \in X$. Then $T$ has a unique fixed point in ${X}$. Moreover, for any $e(0) \in X$ the sequence $e(n)$ defined as $e(n+1)=T(e(n))$,  
converges to the fixed point of $T$.
\label{theo}
\end{theorem}
The key element that guarantees such theoretical asymptotic stability is that the ReNN distance ratio is always below 1. In fact, it is easy to demonstrate that the updating rule of Eq.~\ref{eligibility} is a contraction and converges to its unique fixed point 0 according to the Contraction Mapping theorem (Banach fixed-point theorem).
\\
\indent
The asymptotic stability of the method is illustrated in Fig.~\ref{asymptotic} with a simple one-dimensional case. Two patterns of synthetic descriptors, respectively modeling the case of a distinctive identity (red curve) and a non distinctive identity (black curve) are generated by two distinct 1D Gaussian distributions. 
The learning method was ran for 1000 iterations for three different configurations of the two distributions. The configurations reflect the limit case in which the distinctiveness assumption of the deep descriptors no longer holds. Mismatches might therefore corrupt the distinctive identity.
The blue points represent the eligibility of the distinctive identity. The histogram in yellow represents the distribution of the distinctive identity as incrementally learned by the system. 
The three figures represent distinct cases in which the non distinctive identity is progressively overlapping the distinctive one. The ReNN matching mechanism and the memory control mechanism still keep the learned distinctive identity close to its ground truth pdf.
\section{Quantitative Experiments}
\label{Experiments}
\begin{figure}[t]
\centering
\hspace{-0.8cm}
\includegraphics[width=0.49\textwidth]{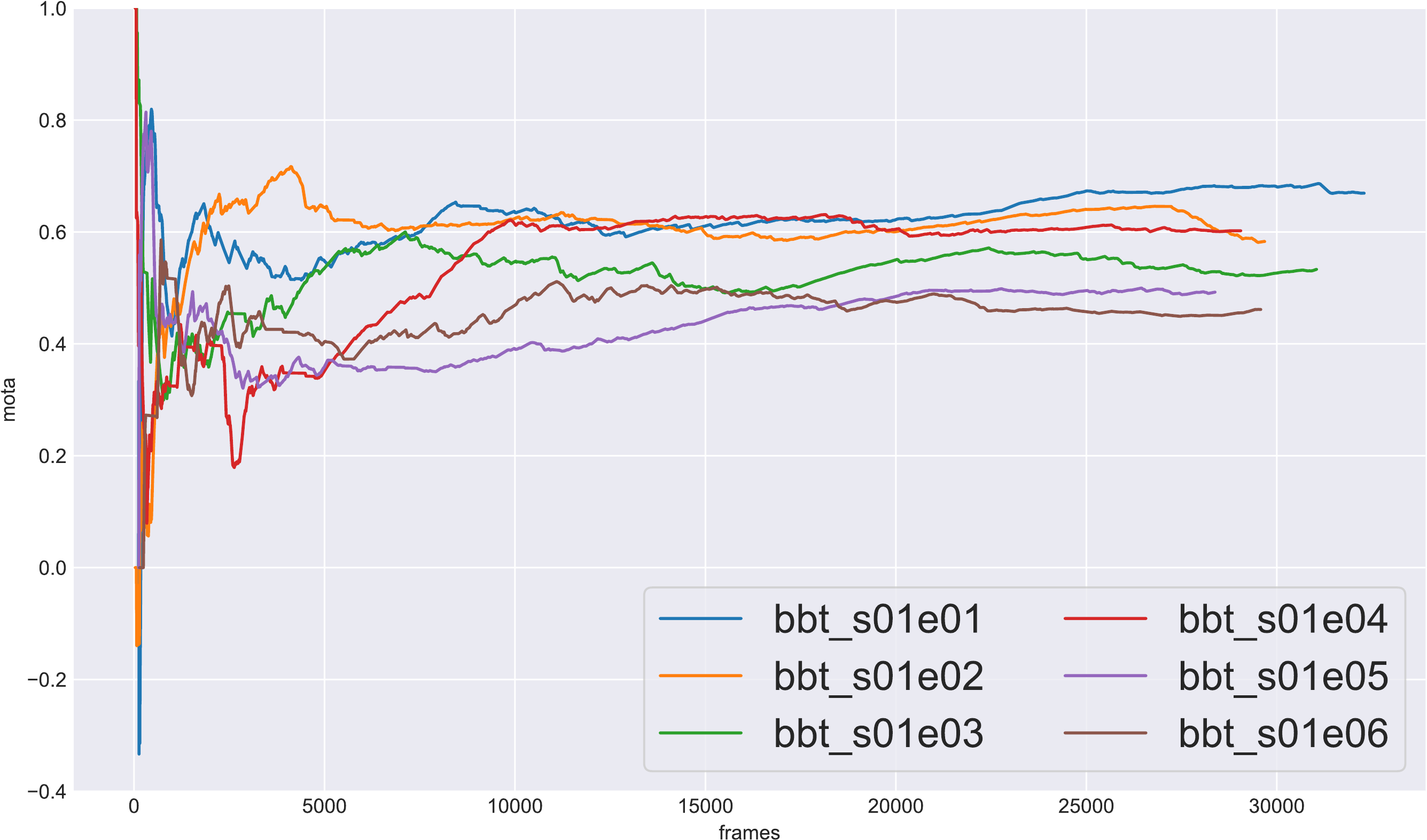}
\caption{MOTA for each video sequence in the \emph{BBT} dataset.\vspace{-0.6cm}}
\label{fig:bbt-mota-nodrift}
\vspace{-0.1cm}
\end{figure}

We focus on tracking/identifying multiple faces according to their unknown identities in unconstrained videos consisting of many shots typically taken from different cameras. We used the \emph{Music}-dataset in \cite{zhang2016tracking} which includes 8 music videos downloaded from YouTube with annotations of 3,845 face tracks and 117,598 face detections. 
We also add the first 6 episodes from Season 1 of the Big Bang Theory TV Sitcom (referred as \emph{BBT}01-06) \cite{wu2013simultaneous}. Each video is about more than 20 minutes long with 5-13 people and is taken mostly indoors. The main difficulty lies in identifying faces of the same subject from a long video footage.
\input{comparisonTable.tex}
\\
\indent
The two algorithm parameters in Eq.~\ref{eq_squaredratio} are set empirically to: $\bar{\rho}=1.6$ and $\alpha=0.01$. Deep face descriptor are extracted according to \cite{Parkhi15}.
We firstly show the capability of the proposed method to perform online learning without drifting using the long sequences of the \emph{BBT} dataset. This consists on monitoring the performance degradation of the system as time advances. A decrease in performance may eventually hinder learning being the system in a condition from which is not possible to recover.
In order to build a picture of the performance over time we evaluate the method with the metric set commonly used in multiple object tracking \cite{leal2015motchallenge}. In particular we report the MOTA: The Multiple Object Tracking Accuracy that takes into account false positives, wrongly rejected identities and identity switches as:
$
	\text{MOTA} = 1 - \frac{\sum_t (\text{FN}_t+\text{FP}_t + \text{IDS}_t) }{\sum_t \text{GT}_t}
    \label{mota}
$
where $\text{GT}_t$, $\text{FN}_t$, $\text{FP}_t$ and $\text{IDS}_t$ are respectively the number of ground truth objects, the number of false negatives, 
the number of false positives and the number of identity switches at time $t$. 
The identity switches are defined as the total number of times that a tracked trajectory changes its matched GT identity. Fig.~\ref{fig:bbt-mota-nodrift} shows the MOTA curves as time progresses for each video sequence of the \emph{BBT} dataset for about 30000 frames. Each individual frame is used to test the model before it is used for training by the incremental learning procedure \cite{ACMsurveyConcept}.
As can be seen from the figure the curves reveal the stability of the learning mechanism confirming the theoretical result of Sec.~\ref{AsymptoticStability}. 
The initial fluctuations typically vary from sequence to sequence and reflect the approximate invariance of the original representation. That is, the few descriptors entering in the memory at the beginning of each sequence do not provide substantial improvement with respect to the original representation. However, as time advances, the reduction of fluctuations reveal that the proposed method is able to learn by collecting all the non-redundant descriptors it can from the video stream until no more improvement is possible.
\\
\indent
We further compare the proposed algorithm with other state-of-the-art MOT trackers, including modified versions of TLD \cite{kalal2012tracking}, ADMM \cite{dicle2013way}, IHTLS \cite{ayazoglu2012fast}. 
We specifically compare with two multi-face tracking methods using the TLD method implemented as described in \cite{zhang2016tracking}. The first method, called mTLD, runs multiple TLD trackers for all targets, and each TLD tracker is initialized with the ground truth bounding box in the first frame. The second method, referred as mTLD2, is used to generate shot-level trajectories
within each shot initializing TLD trackers with untracked detections, and link the detections in the following frames according to their overlap scores with TLD outputs. 
\\
\indent
The methods indicated as Pre-trained, SymTriplet, Triplet and Siamese refers to the four alternatives methods proposed in \cite{zhang2016tracking}.
In these methods including ADMM, mTLD, mTLD2 and IHTLS, shot changes are detected and the video is divided into non-overlapping shots. Within each shot, a face detector is applied and adjacent detections are linked into tracklets. The recovered collection of tracklets are used as face pairs (Siamese) or face triplets (Triplet and SymTriplet) to fine-tune a CNN initial face feature descriptor based on the AlexNet architecture trained on the CASIA-WebFace (Pre-trained). Then, appearance of each detection is represented with the fine-tuned feature descriptors and tracklets within each shot are linked into shot-level tracklets according to a global multiple object tracking \cite{huang2008robust,xing2009multi}. Finally tracklets across shots are subsequently merged across multiple shots into final trajectories according to the Hierarchical Agglomerative Clustering.
\\
\indent
We reported two alternative versions using the (Deformable Part Model) DPM \cite{mathias2014face} and the Tiny \cite{hu2016finding} face detectors. They are  indicated as \textbf{MuFTiR}-\textsc{dpm} and \textbf{MuFTiR}-\textsc{tiny} respectively.
For such comparisons we also include the multiple target metric MOTP: The Multiple Object Tracking Precision. MOTP is the average dissimilarity between all true positives and their corresponding ground truth targets. MOTP is a measure of localization precision.
Given the quite different nature between offline and online this comparison is to be considered a proof-of-concept. However, given the good performance of the offline methods we compare to, it is certainly non-trivial for our online method to do any better. 
Table~\ref{table-compare-1} shows that our online tracking algorithm does reasonably well with respect to offline algorithms, although there are some exceptions.
In \textsc{HelloBubble}, \textsc{BrunoMars}, \textsc{Darling}, \textsc{Tara} and \textsc{WestLife} our best performing method has the MOTA score similar to the ADMM and IHTLS methods with little less identity switches. Despite the on par performance, our method achieves the results without exploiting future information. Performance are still good in \textsc{Apink}, the identity switches are still comparable despite a decrease in MOTA.  
Excluding Siamese, Triplet and SymTriplet that use a refined descriptor specifically tailored to the clustered identities extracted with the multiple passes over the sequence, our method is on par with the other offline methods. Our main observation is that with modern CNN based face detector and descriptor, the state-of-the-art offline trackers do not have expected advantages over the simpler online ones. Advantages further thin when processing long video sequences that do not fit into memory.
\\
\indent
\begin{table*}[h]
	\centering
	\caption{Quantitative comparison with other state-of-the-art multi-object tracking methods on the \emph{BBT} dataset.}
	\label{table-bbt-mota}	
	\begin{minipage}{.33 \linewidth}
   \centering
		\resizebox{1\textwidth}{!}{
			\begin{tabular}{|l|l|r|r|r|}
				\hline
				\multicolumn{5}{|c|}{\textsc{ bbt\_s01e01 }}\\
				\hline
				Method               & Mode    &   IDS $\downarrow$ &   MOTA $\uparrow$ &   MOTP $\uparrow$ \\
				\hline
				mTLD \cite{Kalal-PAMI-2011}                & Offline &                  1 &             -16.3 &              74.8 \\
				ADMM \cite{dicle2013way}                & Offline &                323 &              42.5 &              64.0 \\
				IHTLS \cite{ayazoglu2012fast}               & Offline &                312 &              45.7 &              64.0 \\
				Pre-Trained \cite{zhang2016tracking}         & Offline &                171 &              41.9 &              73.3 \\
				mTLD2 \cite{Kalal-PAMI-2011}               & Offline &                223 &              58.4 &              73.8 \\
				Siamese \cite{zhang2016tracking}             & Offline &                144 &              69.0 &              73.7 \\
				Triplet \cite{zhang2016tracking}             & Offline &                164 &              69.3 &              73.6 \\
				SymTriplet \cite{zhang2016tracking}          & Offline &                156 &              72.2 &              73.7 \\\hline
				\textbf{MuFTiR}-tiny & \textbf{Online}  &                 24 &              59.9 &              70.3 \\
				\hline
		\end{tabular}}
	\end{minipage}%
	\begin{minipage}{.33\linewidth}
		\resizebox{1\textwidth}{!}{
			\begin{tabular}{|l|l|r|r|r|}
				\hline
				\multicolumn{5}{|c|}{\textsc{ bbt\_s01e02 }}\\
				\hline
				Method               & Mode    &   IDS $\downarrow$ &   MOTA $\uparrow$ &   MOTP $\uparrow$ \\
				\hline
				mTLD                 & Offline &                  1 &              -7.6 &              82.8 \\
				ADMM                 & Offline &                395 &              41.3 &              71.3 \\
				IHTLS                & Offline &                394 &              42.4 &              71.4 \\
				Pre-Trained          & Offline &                130 &              27.4 &              74.5 \\
				mTLD2                & Offline &                174 &              43.6 &              75.9 \\
				Siamese              & Offline &                116 &              60.4 &              75.8 \\
				Triplet              & Offline &                143 &              60.2 &              75.7 \\
				SymTriplet           & Offline &                102 &              61.6 &              75.7 \\\hline
\textbf{MuFTiR}-tiny & \textbf{Online}  &                 57 &              45.1 &              68.8 \\
				\hline
		\end{tabular}}
	\end{minipage}
	\begin{minipage}{.33\linewidth}\centering
		\resizebox{1\textwidth}{!}{
			\begin{tabular}{|l|l|r|r|r|}
				\hline
				\multicolumn{5}{|c|}{\textsc{ bbt\_s01e03 }}\\
				\hline
				Method               & Mode    &   IDS $\downarrow$ &   MOTA $\uparrow$ &   MOTP $\uparrow$ \\
				\hline
				mTLD                 & Offline &                  5 &              -2.1 &              69.4 \\
				ADMM                 & Offline &                370 &              30.8 &              68.1 \\
				IHTLS                & Offline &                376 &              33.5 &              68.0 \\
				Pre-Trained          & Offline &                110 &              17.8 &              67.5 \\
				mTLD2                & Offline &                142 &              38.0 &              67.9 \\
				Siamese              & Offline &                109 &              52.6 &              67.9 \\
				Triplet              & Offline &                121 &              50.7 &              67.8 \\
				SymTriplet           & Offline &                126 &              51.9 &              67.8 \\\hline
\textbf{MuFTiR}-tiny & \textbf{Online}  &                 14 &              43.6 &              68.4 \\
				\hline
		\end{tabular}}
	\end{minipage}%
\\
	\begin{minipage}{.33\linewidth}\centering
		\resizebox{1\textwidth}{!}{
			\begin{tabular}{|l|l|r|r|r|}
				\hline
				\multicolumn{5}{|c|}{\textsc{ bbt\_s01e04 }}\\
				\hline
				Method               & Mode    &   IDS $\downarrow$ &   MOTA $\uparrow$ &   MOTP $\uparrow$ \\
				\hline
				mTLD                 & Offline &                  0 &             -15.9 &              76.8 \\
				ADMM                 & Offline &                298 &               9.7 &              65.8 \\
				IHTLS                & Offline &                295 &              13.3 &              65.8 \\
				Pre-Trained          & Offline &                 46 &               0.1 &              66.3 \\
				mTLD2                & Offline &                103 &              11.6 &              66.3 \\
				Siamese              & Offline &                 85 &              23.0 &              66.4 \\
				Triplet              & Offline &                103 &              18.0 &              66.4 \\
				SymTriplet           & Offline &                 77 &              19.5 &              66.4 \\\hline
				\textbf{MuFTiR}-tiny & \textbf{Online}  &                 84 &              53.2 &              69.6 \\
				\hline
		\end{tabular}}
	\end{minipage}
	\begin{minipage}{.33\linewidth}\centering
		\resizebox{1\textwidth}{!}{
			\begin{tabular}{|l|l|r|r|r|}
				\hline
				\multicolumn{5}{|c|}{\textsc{ bbt\_s01e05 }}\\
				\hline
				Method               & Mode    &   IDS $\downarrow$ &   MOTA $\uparrow$ &   MOTP $\uparrow$ \\
				\hline
				mTLD                 & Offline &                  1 &             -15.5 &              76.9 \\
				ADMM                 & Offline &                380 &              37.4 &              68.2 \\
				IHTLS                & Offline &                360 &              33.8 &              68.2 \\
				Pre-Trained          & Offline &                 98 &              32.3 &              75.0 \\
				mTLD2                & Offline &                169 &              46.4 &              74.9 \\
				Siamese              & Offline &                128 &              60.7 &              75.0 \\
				Triplet              & Offline &                118 &              60.5 &              74.9 \\
				SymTriplet           & Offline &                 90 &              60.9 &              74.9 \\\hline
				\textbf{MuFTiR}-tiny & \textbf{Online}  &                 36 &              44.5 &              69.3 \\
				\hline
		\end{tabular}}
	\end{minipage}%
	\begin{minipage}{.33\linewidth}\centering
		\resizebox{1\textwidth}{!}{
			\begin{tabular}{|l|l|r|r|r|}
				\hline
				\multicolumn{5}{|c|}{\textsc{ bbt\_s01e06 }}\\
				\hline
				Method               & Mode    &   IDS $\downarrow$ &   MOTA $\uparrow$ &   MOTP $\uparrow$ \\
				\hline
				mTLD                 & Offline &                  0 &              -3.9 &              89.3 \\
				ADMM                 & Offline &                527 &              47.5 &              97.6 \\
				IHTLS                & Offline &                515 &              43.2 &              97.7 \\
				Pre-Trained          & Offline &                191 &              27.8 &              98.2 \\
				mTLD2                & Offline &                192 &              37.7 &              97.8 \\
				Siamese              & Offline &                156 &              46.2 &              97.9 \\
				Triplet              & Offline &                185 &              45.4 &              98.0 \\
				SymTriplet           & Offline &                196 &              47.6 &              98.0 \\\hline
				\textbf{MuFTiR}-tiny & \textbf{Online}  &                222 &              42.9 &              69.2 \\

				\hline
		\end{tabular}}
	\end{minipage}%
	\vspace{-0.4cm}
\end{table*}
Results are confirmed in the \emph{BBT} dataset as shown in Table~\ref{table-bbt-mota}. As in the previous comparison on the \emph{Music} dataset, except for the Siamese, Triplet and SymTriplet the overall performance are very good. In the Episode four we achieved better results.
Considering that CNN descriptor fine-tuning takes around 1 hour per sequence on a modern GPU, our method perform favorably in those applications operating on real time streaming data. Currently our approach runs at 10 frame per second on 800x600 video frame resolution on a Nvidia GeForce GTX TITAN X (Maxwell). 
\\
\indent
\begin{table}[b] \vspace{-0.4cm}
	\centering
	\caption{Clustering results on \emph{Music} Dataset. Weighted purity of each video is measured on ideal number of clusters.}
	\label{table-music-purity}
	\begin{minipage}{1.02\linewidth}
		\resizebox{1\textwidth}{!}{
		\begin{tabular}{|l|c|c|c|c|c|c|c|c|}
			\hline
			\multicolumn{9}{|c|}{\textsc{ Music Dataset }}\\
			\hline
			Videos                 &        Apink  &   B. Mars  &      Darling  &  Girls A.  & Hello B.  & P. Dolls  &        T-ara  &     Westlife  \\
			\hline
			HOG                    &         0.20  &         0.36  &         0.19  &         0.29  &         0.35  &           0.28  &         0.22  &         0.27  \\
			AlexNet                &         0.22  &         0.36  &         0.18  &         0.30  &         0.31  &           0.31  &         0.25  &         0.37  \\
			Pre-trained            &         0.29  &         0.50  &         0.24  &         0.33  &         0.34  &           0.31  &         0.31  &         0.37  \\
			VGG-Face               &         0.24  &         0.44  &         0.20  &         0.31  &         0.29  &           0.46  &         0.23  &         0.27  \\
			Siamese                &         0.48  &         0.88  &         0.46  &         0.67  &         0.54  &           0.77  &         0.69  &         0.54  \\
			Triplet                &         0.60  &         0.83  &         0.49  &         0.67  &         0.60  &           0.77  &         0.68  &         0.52  \\
			SymTriplet             & \textbf{0.72} &         0.90  &         0.70  &         0.69  & \textbf{0.64} &           0.78  &         0.69  &         0.56  \\ \hline
			\textbf{MuFTiR}-tiny   &         0.51  & \textbf{0.96} & \textbf{0.73} & \textbf{0.89} &         0.59  &   \textbf{0.97} & \textbf{0.72} & \textbf{0.98} \\
			\hline
		\end{tabular}}
\end{minipage}%
\vspace{-0.3cm}
\end{table}
\begin{table}[b]
	\centering
	\caption{Clustering results on Big Bang Theory Dataset. Weighted purity of each video is measured on ideal number of clusters.}
	\label{table-bbt-purity}
	\begin{minipage}{.99\linewidth}\centering
		\resizebox{1\textwidth}{!}{
		\begin{tabular}{|l|c|c|c|c|c|c|}
			\hline
			\multicolumn{7}{|c|}{\textsc{ Big Bang Theory }}\\
			\hline
			Episodes                  &         BBT01 &        BBT02  &        BBT03  &        BBT04  &        BBT05  &        BBT06  \\
			\hline
			HOG                     &         0.37  &         0.32  &         0.38  &         0.35  &         0.29  &         0.26  \\
			AlexNet                 &         0.47  &         0.32  &         0.45  &         0.35  &         0.29  &         0.26  \\
			Pre-trained             &         0.62  &         0.72  &         0.73  &         0.57  &         0.52  &         0.52  \\
			VGG-Face                &         0.91  &         0.85  &         0.83  &         0.54  &         0.65  &         0.46  \\
			Siamese            &         0.94  &         0.95  &         0.87  &         0.74  &         0.70  &         0.70  \\
			Triplet            &         0.94  &         0.95  &         0.92  &         0.74  &         0.68  &         0.70  \\
			SymTriplet         &         0.94  &         0.95  &         0.92  &         0.78  &         0.85  &         0.75  \\ \hline
			\textbf{MuFTiR}-tiny    & \textbf{0.98} & \textbf{0.98} & \textbf{0.98} & \textbf{0.85} & \textbf{0.98} & \textbf{0.94} \\
			\hline
		\end{tabular}}
	\end{minipage}%
\end{table}
MOTA\footnote{Provided by www.motchallenge.org}, while largely used to evaluate performance in multiple object tracking, it is not fully appropriate to evaluate the performance of identification in a \emph{open world} scenario. 
In fact, it does not explicitly handle target re-identification. Different identities assigned to the same individual in two distinct scenes are not accounted as an identity switch. This effect has particular impact with videos obtained from multiple cameras or with many shots. 
In order to take into account this case, for each sequence we also report the weighted cluster purity, defined as: $W =\frac{1}{M} \sum_c  m_c · p_c$, where $c$ is the identity cluster, $m_c$ the number of assigned identities, $p_c$ the ratio between the most frequently occurred identity and $m_c$, and $M$ denotes the total number of identity detections in the video.
Table~\ref{table-music-purity} and \ref{table-bbt-purity} show the quantitative results of the comparison with the \emph{Music} and the \emph{BBT} datasets. HOG, AlexNet and VGGface indicate the method \cite{zhang2016tracking} using alternative descriptors. HOG uses a conventional hand-crafted feature with 4356 dimensions, AlexNet uses a generic feature representation with 4096 dimensions.
Our proposed approach achieves the best performance in six out of eight videos in the \emph{Music} dataset and it achieves state of the art in all the \emph{BBT} video sequences.
\\
\indent
Finally, Fig.~\ref{brunoMarsFrames} shows four frames of the of the \textsc{Bruno Mars} sequence with the learned identities superimposed. Faces appear sensibly diverse (see f.e. individual number 1), nonetheless it can be observed that the learning mechanism is capable to extend the original representation to preserve identities under large pose variations including face profiles not included in the original representation. 
\begin{figure}
\centering
\includegraphics[width=0.23\textwidth]{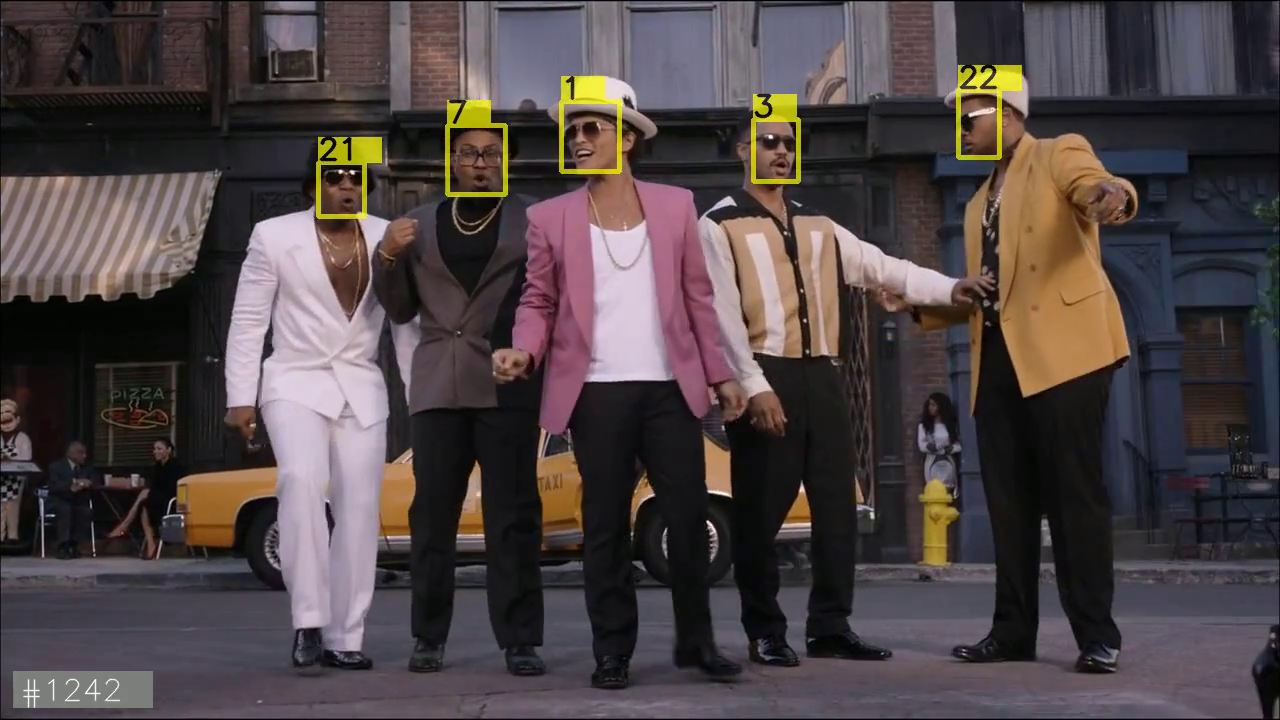}
\includegraphics[width=0.23\textwidth]{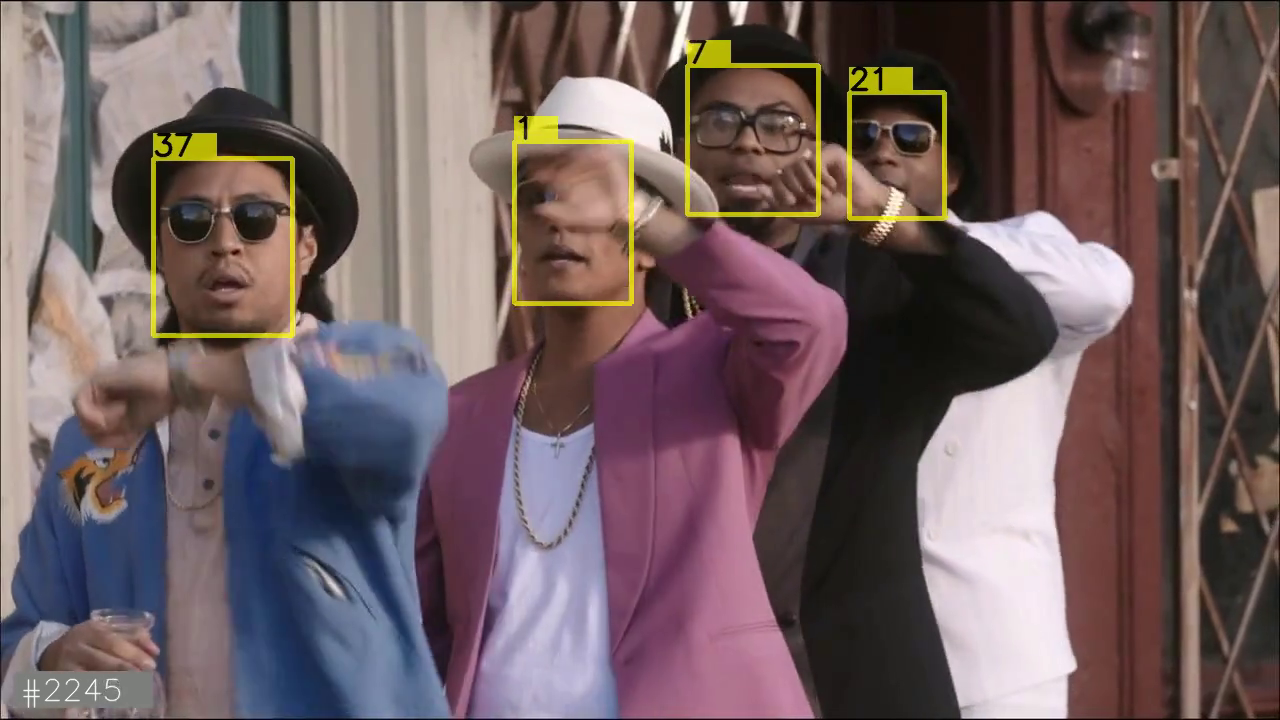}
\includegraphics[width=0.23\textwidth]{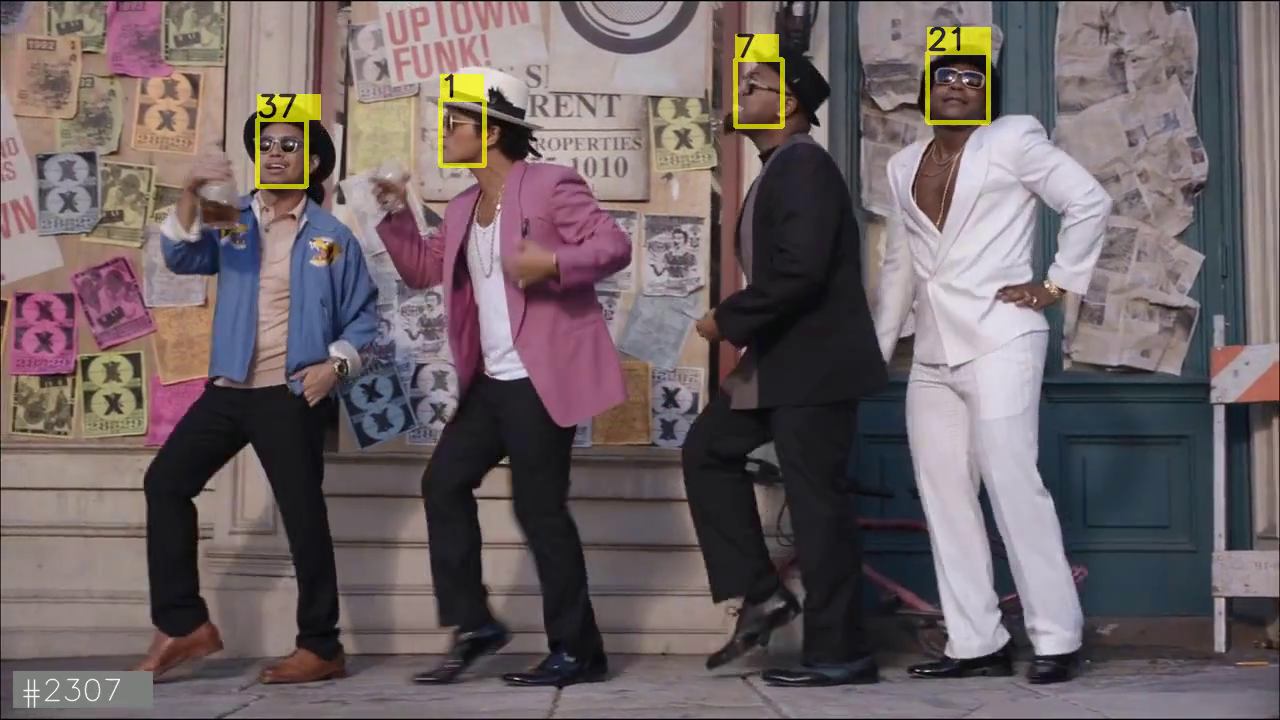}
\includegraphics[width=0.23\textwidth]{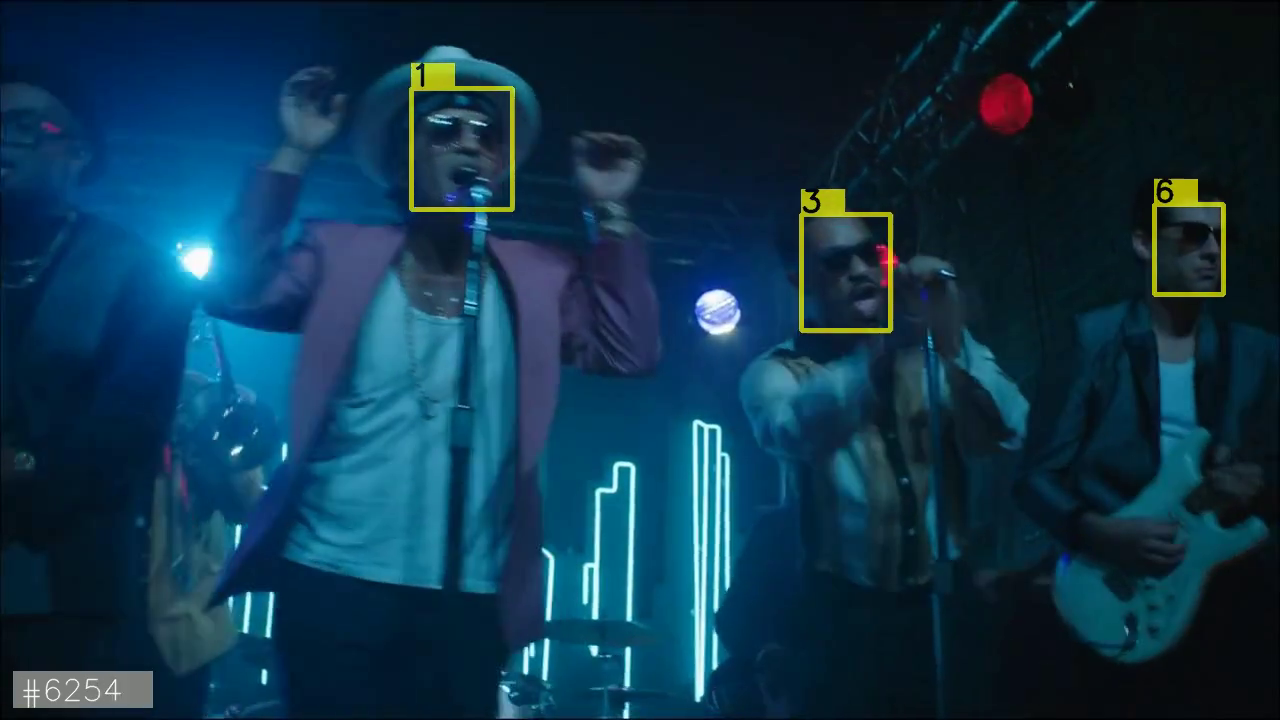}
\caption{Four frames from the \textsc{BrunoMars} video sequence with the superimposed estimated identities are shown. \vspace{-0.5cm}}
\label{brunoMarsFrames}
\end{figure}
\vspace{-0.2cm}
\section{Conclusion}
In this paper we exploited deep CNN based face detection and descriptors coupled with a novel memory based learning mechanism that learns face identities from video sequences unsupervisedly, exploiting the temporal coherence of video frames. Particularly, all the past observed information is learned in a comprehensive representation. 
We demonstrate the effectiveness of the proposed method with respect multiple face tracking on the \emph{Music} and \emph{BBT} datasets.
The proposed method is simple, theoretically sound, asymptotically stable and follows the cumulative and convergent nature of human learning. It can be applied in principle to any other context for which a detector-descriptor combination is available (i.e. car, person, boat, traffic sign).

\section*{Acknowledgment}  This research is based upon work supported in part by the Office of the Director of National Intelligence (ODNI), Intelligence Advanced Research Projects Activity (IARPA), via IARPA contract number 2014-14071600011. The views and conclusions contained herein are those of the authors and should not be interpreted as necessarily representing the official policies or endorsements, either expressed or implied, of ODNI, IARPA, or the U.S. Government. The U.S. Government is authorized to reproduce and distribute reprints for Governmental purpose notwithstanding any copyright annotation thereon.

{\small
\bibliographystyle{unsrt}
\bibliography{IEEEfullMy}
}

\end{document}

%% file: comparisonTable.tex
\begin{table*}[t]
	\caption{Quantitative comparison with other state-of-the-art multi-object tracking methods on the \emph{Music} video dataset}
	\label{table-compare-1}
	\centering
	\begin{minipage}{.33\linewidth}
		\resizebox{1\textwidth}{!}{
			\begin{tabular}{|l|l|r|r|r|}
				\hline
				\multicolumn{5}{|c|}{\textsc{ Apink }}\\
				\hline
				Method               & Mode    &   IDS $\downarrow$ &   MOTA $\uparrow$ &   MOTP $\uparrow$ \\
				\hline
				mTLD \cite{Kalal-PAMI-2011}                & Offline &                 31 &              -2.2 &              71.2 \\
				ADMM \cite{dicle2013way}                 & Offline &                179 &              72.4 &              76.1 \\
				IHTLS \cite{ayazoglu2012fast}                & Offline &                173 &              74.9 &              76.1 \\
				Pre-Trained \cite{zhang2016tracking}         & Offline &                100 &              54.0 &              75.5 \\
				mTLD2 \cite{Kalal-PAMI-2011}               & Offline &                173 &              77.4 &              76.3 \\
				Siamese \cite{zhang2016tracking}             & Offline &                124 &              79.0 &              76.3 \\
				Triplet \cite{zhang2016tracking}             & Offline &                140 &              78.9 &              76.3 \\
				SymTriplet \cite{zhang2016tracking}          & Offline &                 78 &              80.0 &              76.3 \\ 							\hline
				\textbf{MuFTiR}-dpm  & \bf{Online}  &                121 &              21.8 &              61   \\
				\textbf{MuFTiR}-tiny & \bf{Online}  &                191 &              55.1 &              65.4 \\
				\hline
		\end{tabular}}
	\end{minipage}%
	\begin{minipage}{.33\linewidth}
		\resizebox{1\textwidth}{!}{
			\begin{tabular}{|l|l|r|r|r|}
				\hline
				\multicolumn{5}{|c|}{\textsc{ BrunoMars }}\\
				\hline
				Method               & Mode    &   IDS $\downarrow$ &   MOTA $\uparrow$ &   MOTP $\uparrow$ \\
				\hline
				mTLD                 & Offline &                 35 &              -8.7 &              65.3 \\
				ADMM                 & Offline &                428 &              50.6 &              85.7 \\
				IHTLS                & Offline &                375 &              52.7 &              85.8 \\
				Pre-Trained          & Offline &                151 &              48.3 &              88.0 \\
				mTLD2                & Offline &                278 &              52.6 &              87.9 \\
				Siamese              & Offline &                126 &              56.7 &              87.8 \\
				Triplet              & Offline &                126 &              56.6 &              87.8 \\
				SymTriplet           & Offline &                105 &              56.8 &              87.8 \\ \hline
				\textbf{MuFTiR}-dpm  & \bf{Online}  &                78  &               4.5 &              61   \\
				\textbf{MuFTiR}-tiny & \bf{Online}  &                420 &              48.8 &              65.5 \\
				\hline
		\end{tabular}}
	\end{minipage}
	\begin{minipage}{.33\linewidth}
		\resizebox{1\textwidth}{!}{
			\begin{tabular}{|l|l|r|r|r|}
				\hline
				\multicolumn{5}{|c|}{\textsc{ Darling }}\\
				\hline
				Method               & Mode    &   IDS $\downarrow$ &   MOTA $\uparrow$ &   MOTP $\uparrow$ \\
				\hline
				mTLD                 & Offline &                 24 &             -22.0 &              69.9 \\
				ADMM                 & Offline &                412 &              53.0 &              88.4 \\
				IHTLS                & Offline &                381 &              62.7 &              88.4 \\
				Pre-Trained          & Offline &                115 &              42.7 &              88.5 \\
				mTLD2                & Offline &                278 &              59.8 &              89.3 \\
				Siamese              & Offline &                214 &              69.5 &              88.9 \\
				Triplet              & Offline &                187 &              69.2 &              88.9 \\
				SymTriplet           & Offline &                169 &              70.5 &              88.9 \\ \hline
				\textbf{MuFTiR}-dpm  & \bf{Online}  &                64  &              2.2  &              63.7 \\
				\textbf{MuFTiR}-tiny & \bf{Online}  &                449 &              62.1 &              66.0 \\
				\hline
		\end{tabular}}
	\end{minipage}%
\\	
	\begin{minipage}{.33\linewidth}
		\resizebox{1\textwidth}{!}{
			\begin{tabular}{|l|l|r|r|r|}
				\hline
				\multicolumn{5}{|c|}{\textsc{ GirlsAloud }}\\
				\hline
				Method               & Mode    &   IDS $\downarrow$ &   MOTA $\uparrow$ &   MOTP $\uparrow$ \\
				\hline
				mTLD                 & Offline &                  9 &              -1.1 &              71.0 \\
				ADMM                 & Offline &                487 &              46.6 &              87.1 \\
				IHTLS                & Offline &                396 &              51.8 &              87.2 \\
				Pre-Trained          & Offline &                138 &              42.7 &              87.7 \\
				mTLD2                & Offline &                322 &              46.7 &              88.2 \\
				Siamese              & Offline &                112 &              51.6 &              87.8 \\
				Triplet              & Offline &                 80 &              51.7 &              87.8 \\
				SymTriplet           & Offline &                 64 &              51.6 &              87.8 \\ \hline
				\textbf{MuFTiR}-dpm  & \bf{Online}  &                 51 &              -2.7 &              61   \\
				\textbf{MuFTiR}-tiny & \bf{Online}  &                339 &              49.3 &              66.1 \\
				\hline
		\end{tabular}}
	\end{minipage}
	\begin{minipage}{.33\linewidth}
		\resizebox{1\textwidth}{!}{
			\begin{tabular}{|l|l|r|r|r|}
				\hline
				\multicolumn{5}{|c|}{\textsc{ HelloBubble }}\\
				\hline
				Method               & Mode    &   IDS $\downarrow$ &   MOTA $\uparrow$ &   MOTP $\uparrow$ \\
				\hline
				mTLD                 & Offline &                  7 &              -3.5 &              66.5 \\
				ADMM                 & Offline &                115 &              47.6 &              69.9 \\
				IHTLS                & Offline &                109 &              52.0 &              69.9 \\
				Pre-Trained          & Offline &                 71 &              36.6 &              68.5 \\
				mTLD2                & Offline &                139 &              52.6 &              70.5 \\
				Siamese              & Offline &                105 &              56.3 &              70.6 \\
				Triplet              & Offline &                 82 &              56.2 &              70.5 \\
				SymTriplet           & Offline &                 69 &              56.5 &              70.5 \\ \hline
				\textbf{MuFTiR}-dpm  & \bf{Online}  &                170 &               4.0 &              59.0 \\
				\textbf{MuFTiR}-tiny & \bf{Online}  &                 88 &              51.4 &              69.9 \\
				\hline
		\end{tabular}}
	\end{minipage}%
	\begin{minipage}{.33\linewidth}
		\resizebox{1\textwidth}{!}{
			\begin{tabular}{|l|l|r|r|r|}
				\hline
				\multicolumn{5}{|c|}{\textsc{ PussycatDolls }}\\
				\hline
				Method               & Mode    &   IDS $\downarrow$ &   MOTA $\uparrow$ &   MOTP $\uparrow$ \\
				\hline
				mTLD                 & Offline &                 24 &               3.1 &              71.3 \\
				ADMM                 & Offline &                287 &              63.2 &              63.5 \\
				IHTLS                & Offline &                248 &              70.3 &              63.5 \\
				Pre-Trained          & Offline &                128 &              65.1 &              64.9 \\
				mTLD2                & Offline &                296 &              68.3 &              64.9 \\
				Siamese              & Offline &                107 &              70.3 &              64.9 \\
				Triplet              & Offline &                 99 &              69.9 &              64.9 \\
				SymTriplet           & Offline &                 82 &              70.2 &              64.9 \\ \hline
				\textbf{MuFTiR}-dpm  & \bf{Online}  &                 55 &             -13.5 &              61.1 \\
				\textbf{MuFTiR}-tiny & \bf{Online}  &            	  83 &              30.7 &              62.7 \\

				\hline
		\end{tabular}}
	\end{minipage}
	\begin{minipage}{.33\linewidth}
		\resizebox{1\textwidth}{!}{
			\begin{tabular}{|l|l|r|r|r|}
				\hline
				\multicolumn{5}{|c|}{\textsc{ Tara }}\\
				\hline
				Method               & Mode    &   IDS $\downarrow$ &   MOTA $\uparrow$ &   MOTP $\uparrow$ \\
				\hline
				mTLD                 & Offline &                130 &               1.4 &              67.9 \\
				ADMM                 & Offline &                251 &              29.4 &              63.8 \\
				IHTLS                & Offline &                218 &              35.3 &              63.8 \\
				Pre-Trained          & Offline &                143 &              57.3 &              72.4 \\
				mTLD2                & Offline &                251 &              56.0 &              72.6 \\
				Siamese              & Offline &                106 &              58.4 &              72.5 \\
				Triplet              & Offline &                 94 &              59.0 &              72.5 \\
				SymTriplet           & Offline &                 75 &              59.2 &              72.4 \\ \hline
				\textbf{MuFTiR}-dpm  & \bf{Online}  &                124 &              15   &              68   \\
				\textbf{MuFTiR}-tiny & \bf{Online}  &                270 &              39.5 &              76.4 \\
				\hline
		\end{tabular}}
	\end{minipage}%
	\begin{minipage}{.33\linewidth}
		\resizebox{1\textwidth}{!}{
			\begin{tabular}{|l|l|r|r|r|}
				\hline
				\multicolumn{5}{|c|}{\textsc{ Westlife }}\\
				\hline
				Method               & Mode    &   IDS $\downarrow$ &   MOTA $\uparrow$ &   MOTP $\uparrow$ \\
				\hline
				mTLD                 & Offline &                 20 &             -34.7 &              56.9 \\
				ADMM                 & Offline &                223 &              62.4 &              87.5 \\
				IHTLS                & Offline &                113 &              60.9 &              87.5 \\
				Pre-Trained          & Offline &                 85 &              57.0 &              88.2 \\
				mTLD2                & Offline &                177 &              58.1 &              88.1 \\
				Siamese              & Offline &                 74 &              64.1 &              88.0 \\
				Triplet              & Offline &                 89 &              64.5 &              88.0 \\
				SymTriplet           & Offline &                 57 &              68.6 &              88.1 \\ \hline
				\textbf{MuFTiR}-dpm  & \bf{Online}  &                 47 &              -0.2 &              61.5 \\
				\textbf{MuFTiR}-tiny & \bf{Online}  &            76 &              58.9 &              66.1 \\
				\hline
		\end{tabular}}
	\end{minipage}
	\vspace{-0.4cm}
\end{table*}